\documentclass{article}

\usepackage{microtype}
\usepackage{graphicx}
\usepackage{subfigure}
\usepackage{booktabs} 

\usepackage{hyperref}
\usepackage{xspace}
\usepackage{spverbatim} 
\usepackage{listings}
\usepackage{xcolor} 
\usepackage[most]{tcolorbox}

\newcommand{\agentrole}[1]{\textbf{\textcolor{purple}{[HarmonyCell]:}} #1}
\lstdefinelanguage{json}{
    basicstyle=\normalfont\ttfamily,
    numberstyle=\scriptsize,
    stepnumber=1,
    numbersep=8pt,
    showstringspaces=false,
    breaklines=true,
    frame=lines,
    backgroundcolor=\color{white},
    literate=
     *{0}{{{\color{blue}0}}}{1}
      {1}{{{\color{blue}1}}}{1}
      {2}{{{\color{blue}2}}}{1}
      {3}{{{\color{blue}3}}}{1}
      {4}{{{\color{blue}4}}}{1}
      {5}{{{\color{blue}5}}}{1}
      {6}{{{\color{blue}6}}}{1}
      {7}{{{\color{blue}7}}}{1}
      {8}{{{\color{blue}8}}}{1}
      {9}{{{\color{blue}9}}}{1}
      {:}{{{\color{red}{:}}}}{1}
      {,}{{{\color{red}{,}}}}{1}
      {\{}{{{\color{black}{\{}}}}{1}
      {\}}{{{\color{black}{\}}}}}{1}
      {[}{{{\color{black}{[}}}}{1}
      {]}{{{\color{black}{]}}}}{1},
}

\newtcolorbox{promptbox}[1][]{
    colback=gray!5!white,
    colframe=gray!75!black,
    fonttitle=\bfseries,
    coltitle=white,
    left=5pt, right=5pt, top=2pt, bottom=2pt,
    arc=2pt,
    boxrule=0.5pt,
    enhanced,
    breakable,
    attach boxed title to top left={xshift=3mm, yshift=-2mm, yshifttext=-1mm},
    boxed title style={colback=gray!75!black},
    title=#1
}

\newcommand{\agentname}{HarmonyCell\xspace}

\usepackage[accepted]{icml2026}

\usepackage{amsmath}
\usepackage{amssymb}
\usepackage{mathtools}
\usepackage{amsthm}
\usepackage{colortbl}

\usepackage[capitalize,noabbrev]{cleveref}

\theoremstyle{plain}

\theoremstyle{definition}

\theoremstyle{remark}

\usepackage{multirow}      

\usepackage[textsize=tiny]{todonotes}

\newcommand{\gcheck}{\textcolor[rgb]{0, 0.6, 0}{\boldmath\checkmark}}

\icmltitlerunning{\agentname: Automating Single-Cell Perturbation Modeling under Semantic and Distribution Shifts}

\begin{document}

\twocolumn[
\icmltitle{\agentname: Automating Single-Cell Perturbation Modeling under Semantic and Distribution Shifts}



\icmlsetsymbol{equal}{*}

\begin{icmlauthorlist}
\icmlauthor{Wenxuan Huang}{equal,fudaniics,ailab}
\icmlauthor{Mingyu Tsoi}{equal,hkust}
\icmlauthor{Yanhao Huang}{equal,xjlvp}
\icmlauthor{Xinjie Mao}{equal,ailab,iis}
\icmlauthor{Xue XIA}{hkustguang}
\icmlauthor{Hao Wu}{fudaniics,ailab}
\icmlauthor{Jiaqi Wei}{ailab}
\icmlauthor{Yuejin Yang}{fudaniics,ailab}
\icmlauthor{Lang Yu}{ailab}
\icmlauthor{Cheng Tan}{ailab}
\icmlauthor{Xiang Zhang}{ubc}
\icmlauthor{Zhangyang Gao}{ailab}
\icmlauthor{Siqi Sun}{fudaniics,ailab}
\end{icmlauthorlist}

\icmlaffiliation{fudaniics}{Fudan University}
\icmlaffiliation{ailab}{Shanghai Artificial Intelligence Laboratory}
\icmlaffiliation{hkust}{The Hong Kong University of Science and Technology}
\icmlaffiliation{hkustguang}{The Hong Kong University of Science and Technology (Guangzhou)}
\icmlaffiliation{xjlvp}{Xi'an Jiaotong-Liverpool University}
\icmlaffiliation{iis}{Shanghai Innovation Institute}
\icmlaffiliation{ubc}{University of British Columbia}

\icmlcorrespondingauthor{Zhangyang Gao}{gaozhangyang@ailab.org.cn}
\icmlcorrespondingauthor{Siqi Sun}{siqisun@fudan.edu.cn}

\icmlkeywords{Machine Learning, ICML}

\vskip 0.3in
]



\printAffiliationsAndNotice{\icmlEqualContribution} 

\begin{abstract}

Single-cell perturbation studies face dual heterogeneity bottlenecks: (i) semantic heterogeneity—identical biological concepts encoded under incompatible metadata schemas across datasets; and (ii) statistical heterogeneity—distribution shifts from biological variation demanding dataset-specific inductive biases. We propose \agentname, an end-to-end agent framework resolving each challenge through a dedicated mechanism: an LLM-driven Semantic Unifier autonomously maps disparate metadata into a canonical interface without manual intervention; and an \textit{adaptive} Monte Carlo Tree Search engine operates over a \textit{hierarchical action space} to synthesize architectures with optimal statistical inductive biases for distribution shifts. Evaluated across diverse perturbation tasks under both semantic and distribution shifts, \agentname achieves a 95\% valid execution rate on heterogeneous input datasets (versus 0\% for general agents) while matching or even exceeding expert-designed baselines in rigorous out-of-distribution evaluations. This dual-track orchestration enables scalable automatic virtual cell modeling without dataset-specific engineering \footnote{Code is available upon request, which will be released soon.} .


\end{abstract}

\section{Introduction}

\begin{figure}[t]
\begin{center}
\centerline{\includegraphics[width=1.03\columnwidth]{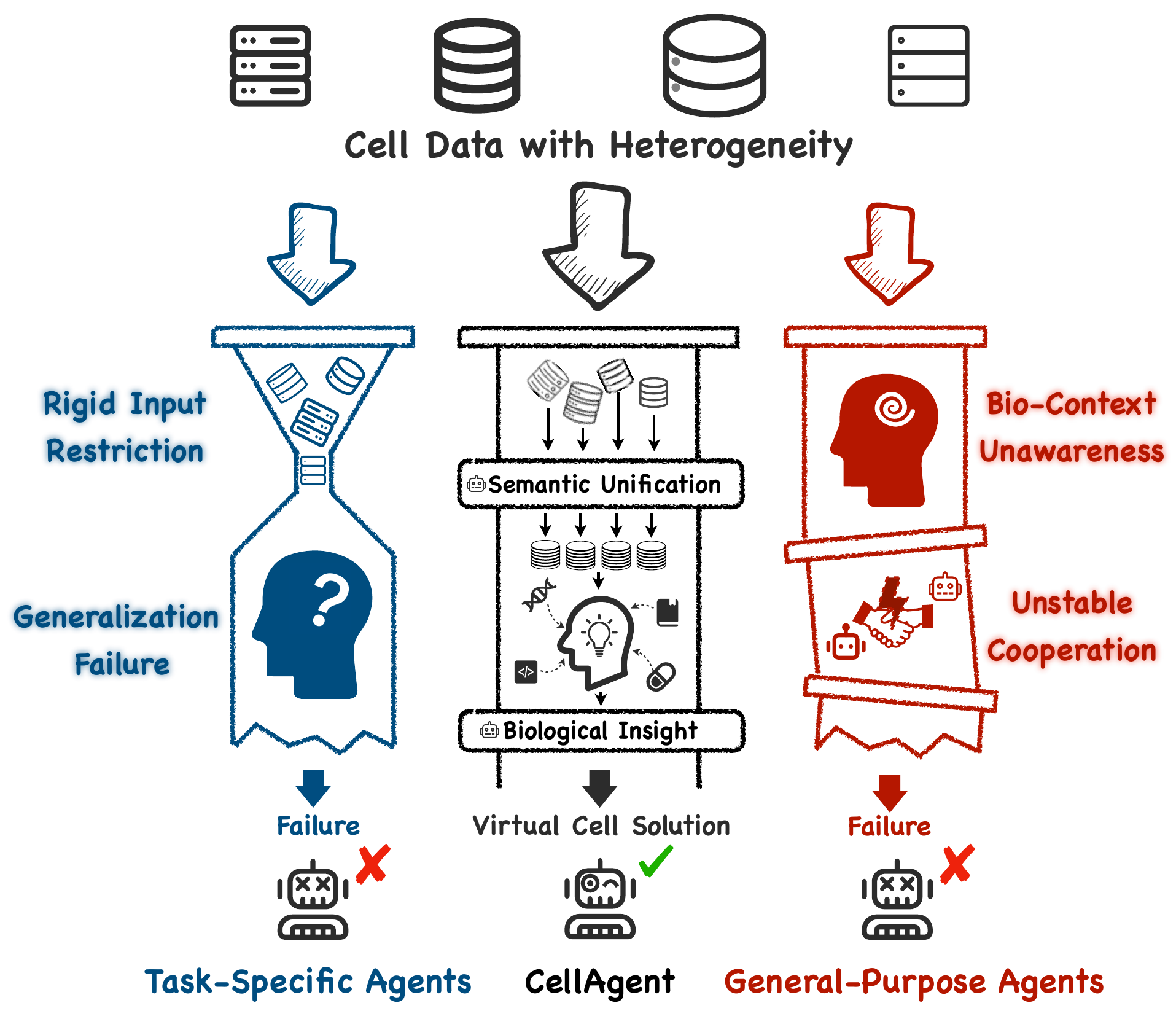}}
\caption{\textbf{Uniqueness of \agentname}. Existing task-specific LLM agents (CellForge etc.) but require rigid data input format, while general-purpose agents often lack biological knowledge. \agentname can enhance its capabilities through biological priors, and also handle the problem of data heterogeneity.}
\label{fig:uniqueness}
\end{center}
\vskip -0.3in
\end{figure}

Single-cell perturbation studies are scaling rapidly across laboratories and experimental platforms~\cite{nawy2014single}, bringing the ``Virtual Cell'' vision closer to reality and creating an urgent need to quickly validate whether a newly generated dataset is worth deeper investment~\cite{bunne2024build}. In practice, however, time-to-insight is still bottlenecked by labor-intensive data curation and the complexity of targeted model design. The core challenge stems from \textbf{dual heterogeneity}: semantic and statistical. \textit{Semantic heterogeneity} arises when the same biological concept is encoded under incompatible metadata schemas, naming conventions, indexing protocols, or preprocessing assumptions---forcing repeated format reconciliation before any model can be trained. \textit{Statistical heterogeneity} arises from real biological variation across tissues, donors, and conditions, inducing distribution shifts that require robust cross-domain transfer. Moreover, strong performance depends not only on input standardization but also on dataset-specific choices of architectures, hyper-parameters, objectives, and domain biases---forcing researchers to repeatedly redesign or search for robust strategies for each new dataset. \textit{How to streamline the ``data unification--model design--evaluation'' workflow for rapid data-value assessment} remains an open question.

Existing approaches still fall short of the end-to-end goal: turning a user's scientific intent into a reusable workflow that adapts to heterogeneous single-cell data. As shown in Fig.~\ref{fig:uniqueness}, search-centric deep research agents~\cite{hu2025flowsearch,chai2025scimaster} can retrieve relevant papers and heuristics, but rarely compile them into robust, executable pipelines that ingest messy single-cell data and reliably produce models. General code-centric AutoML agents~\cite{du2025automlgen,jiang2025aide, ou2025automind} often lack biological priors and degrade into reactive trial-and-error, treating each request as a one-off coding task rather than a structured research process. Closest to our setting, CellForge~\cite{tang2025cellforge} shows that multi-agents can design and implement virtual-cell models from raw data and task descriptions; however, it assumes an over-simplified setting and fails to address the \textbf{dual heterogeneity} challenge. Specifically, it lacks mechanisms to (i) automatically harmonize divergent metadata schemas into a canonical interface, and (ii) systematically optimize structural inductive biases to handle complex distribution shifts (e.g., OOD tasks). Consequently, researchers still lack an agent that can autonomously bridge the gap between raw, heterogeneous data and robust model deployment in the fragmented single-cell ecosystem.

To address these gaps, we propose \agentname, an end-to-end agent that treats virtual-cell modeling as a reusable, shift-aware workflow rather than a dataset-specific one-off script. \agentname systematically resolves the \textbf{dual heterogeneity} through two synergistic components: (i) an \textbf{LLM-driven Semantic Unifier}, which aligns heterogeneous schemas and preprocessing conventions into a canonical interface without manual intervention; and (ii) an \textbf{Adaptive MCTS Engine}, which navigates a hierarchical action space to autonomously synthesize model architectures with optimal statistical inductive biases. By integrating semantic alignment with structural search, \agentname ensures stable, reproducible execution, enabling continual dataset onboarding and robust modeling at scale across the fragmented single-cell ecosystem.

We conduct extensive experiments to evaluate \agentname in both \textbf{single-dataset} and \textbf{multi-dataset} settings. In the single-dataset setting, we assess predictive generalization in two primary and representative tasks: gene and drug perturbation prediction. In the multi-dataset setting, we explicitly stress both semantic shifts and distribution shifts by composing datasets that differ in metadata schemas and preprocessing conventions. Beyond predictive accuracy, we quantify end-to-end deployability through pipeline success rate, as well as practical efficiency and execution reliability. Results demonstrate that \agentname not only automates the labor-intensive unification process but also discovers architectures that match or outperform human-designed baselines in rigorous out-of-distribution evaluations.

Our contributions are summarized as follows:

\textbf{1. Semantic Heterogeneity Solver.} \agentname employs an LLM-driven \textbf{Semantic Unifier} to map disparate metadata into a canonical interface, enabling zero-shot adaptation to uncurated datasets without manual engineering.

\textbf{2. Statistical Heterogeneity Solver.} \agentname utilizes an adaptive \textbf{MCTS Engine} within a \textbf{hierarchical action space}. It dynamically synthesizes architectures tailored to biological distribution shifts, robustly handling both familiar and novel contexts through structured exploration.

\textbf{Comprehensive Empirical Validation.}
\agentname achieves a 95\% pipeline success rate on heterogeneous inputs (vs. 0\% for general agents) and matches expert-level performance in OOD tasks, validating its end-to-end reliability in resolving dual heterogeneity for scalable virtual cell modeling without human intervention.

\begin{figure*}[ht]
    \vskip 0.15in
    \begin{center}
    \centerline{\includegraphics[width=0.95\linewidth]{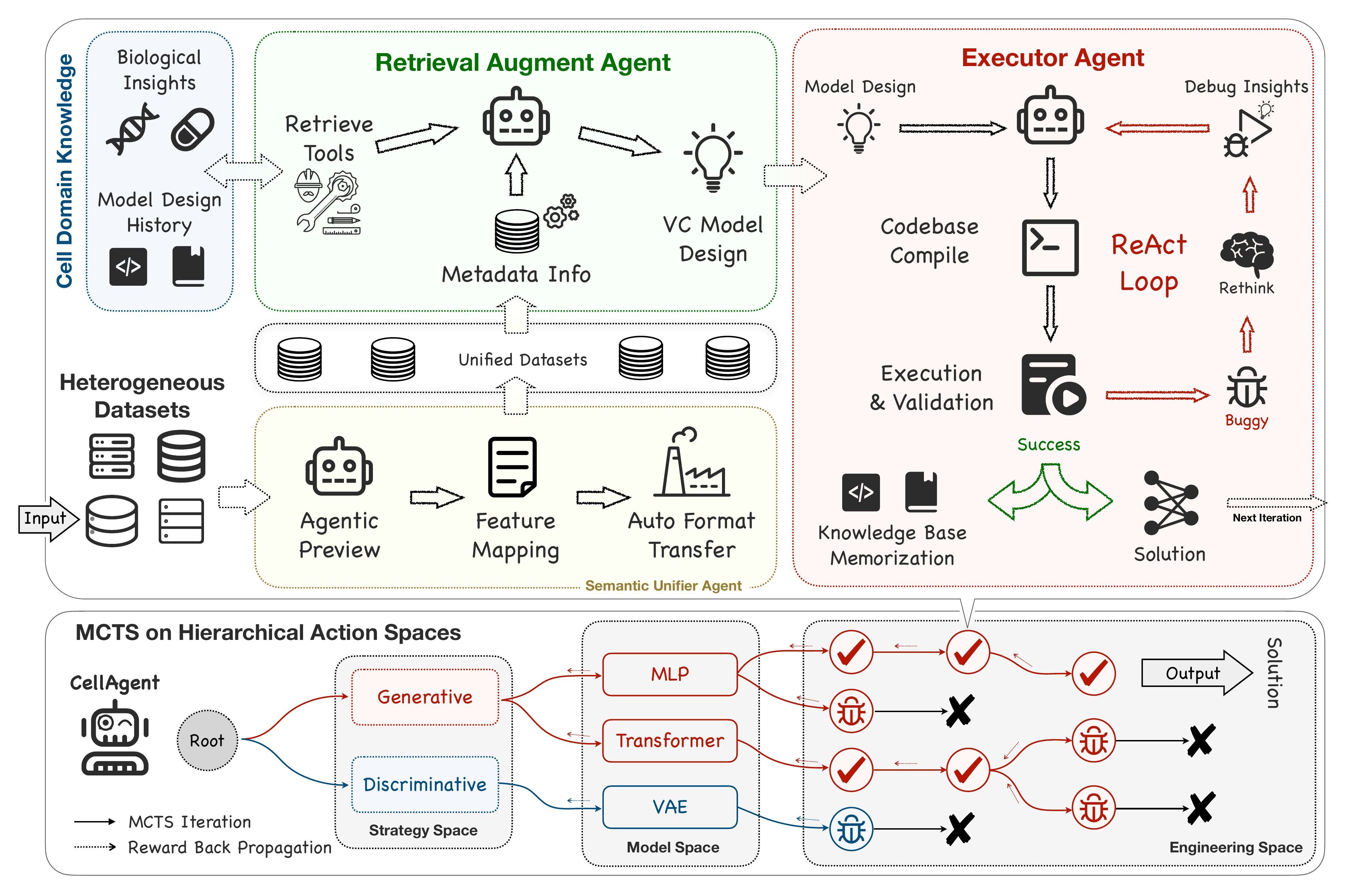}}
    \caption{\textbf{Architecture of \agentname.} The system integrates: (1) an \textbf{LLM-driven Semantic Unifier} to canonicalize heterogeneous \texttt{h5ad} inputs; (2) a \textbf{Retrieval-Augmented Agent} for meta-initialization using historical strategies; and (3) an \textbf{Executor Agent} guided by \textbf{MCTS over a hierarchical action space} (bottom panel). This hierarchical decomposition significantly bolsters search stability by proactively pruning branches that lead to erroneous `bug' nodes before full execution. While runtime failures trigger a ReAct-style debugging loop, successfully validated pipelines are stored in a persistent knowledge base for future reuse.}
    \label{icml-historical}
    \end{center}
    \vskip -0.15in
    \end{figure*}

\section{Related Work}
\subsection{Virtual Cell Modeling in Computational Biology}
Recent large-scale benchmarking efforts have substantially expanded the accessibility of high-quality single-cell perturbation datasets, providing unified evaluation protocols and lowering the barrier to method development \cite{wu2024perturbench, peidli2024scperturb}. Leveraging these resources, deep learning models have rapidly advanced the modeling of cellular dynamics—focusing on learning compact, low-dimensional representations of gene expression to power covariate alignment, perturbation prediction, and multi-modal data integration across heterogeneous experiments \cite{lotfollahi2019scgen, piran2024disentanglement, roohani2024predicting, he2025squidiff, lotfollahi2021compositional}.

\subsection{Coding and Automated Machine Learning Agents}
The intersection of Large Language Models (LLMs) and software engineering has facilitated the development of autonomous agents capable of executing end-to-end machine learning workflows \cite{trirat2024automl, wang2024openhands}. Building upon traditional Automated Machine Learning (AutoML) techniques, recent multi-agent systems leverage the search \cite{hu2025flowsearch, chai2025scimaster} and code-generation \cite{zhuge2024gptswarm, wang2025repomaster, yang2025r, ou2025automind} capabilities of LLMs to automate pipeline construction, ranging from data preprocessing and feature engineering to model architecture search. Recently, some multi-agent systems enables tree-search methods to explore possible solution spaces \cite{jiang2025aide, du2025automlgen}, which accelerate the discovery of scientific problems and coding solutions.

\subsection{AI Scientist: Agents in Scientific Discovery}
Beyond code execution, a distinct class of agents, often termed "AI Scientists," has been developed recently to automate high-level cognitive tasks inherent to the scientific process both in general-purpose domain \cite{wei2025ai, chai2025scimaster, yamada2025ai, lu2024ai, ruan2024automatic, xu2025probing} and specific fields~\cite{ruan2024automatic, jin2025stella, ni2024matpilot, zhang2025origene, gottweis2025towards, swanson2024virtual, xiao2024cellagent}. By using various methods like tree-structure searching\cite{wei2025unifying}, RAG methods\cite{weiretrieval} and data contruction \cite{hu2025survey}, these AI scientists integrate literature retrieval, knowledge extraction, hypothesis generation and code execution to assist researchers in navigating vast scientific corpora and formulating theoretical insights. In agentic virtual cell modeling, CellForge \cite{tang2025cellforge} first merge search and code-generation capabilities of LLMs to automate the design of virtual cell models. 


\section{Preliminaries}

\textbf{Task Formulation.} 
Given a dataset $\mathcal{D} = \{(x^{(i)}, p_i, c_i)\}_{i=1}^N$ with gene expression $x^{(i)} \in \mathbb{R}^G$, perturbation $p$, and covariate $c$, we aim to predict the counterfactual distribution of cells under perturbation $p$. The observed state is modeled as $x_p = x_0 + T_p(x_0, p_i, c_i) + \epsilon$, where $x_0$ is the basal state and $T_p$ is the effect. Since $x_0$ is unobservable, the task is to learn a mapping $f_\theta$ (parameterized by neural networks) that estimates the population-level response $\hat{x}_p$ by minimizing the discrepancy between the predicted and ground-truth distributions in a latent space $z = g_\phi(x)$.

\textbf{Metrics.} 
We evaluate predictions at the pseudo-bulk level. Let $\mathbf{y}$ and $\hat{\mathbf{y}}$ be the true and predicted mean expression vectors. We report: (1) \textbf{RMSE} for global error; (2) \textbf{Pearson Correlation (DeltaPCC)} of the shift vectors ($\delta = \mathbf{y} - \mathbf{y}_{\text{ctrl}}$ vs. $\hat{\delta} = \hat{\mathbf{y}} - \mathbf{y}_{\text{ctrl}}$) to assess linearity; and (3) \textbf{Cosine Similarity (CosLogFC)} of $\delta$ to evaluate directional fidelity. The detailed calculation protocol is provided in Appendix~\ref{supp:metrics}.

\section{Method}
\agentname orchestrates a unified framework to resolve semantic and statistical heterogeneities in virtual cell modeling through two synergistic components: (1) an LLM-driven \textbf{Semantic Unifier} that canonicalizes raw metadata to resolve schema inconsistencies, and (2) an \textbf{Adaptive Monte Carlo Tree Search (MCTS) Engine} operating on a \textbf{Hierarchical Action Space} to systematically optimize statistical inductive biases. This pipeline enables fully automated, end-to-end and stable modeling without manual intervention. Detailed methods is demonstrated in \ref{app:implementation}.

\begin{table*}[ht]
    \centering
    \small
    \begin{tabular}{lcccc}
        \toprule
        \textsc{Abilities} & \textbf{Heterogeneity Data Unification} & \textbf{Biological Prior} & \textbf{Model Exploration} & \textbf{Collaborative Coding} \\
        \midrule
        General-Purpose Agents & $-$ & $-$ & $-$ & \gcheck \\
        Specialized Cell Scientists & $-$ & \gcheck & \gcheck & \gcheck \\
        \textbf{\agentname} & \gcheck & \gcheck & \gcheck & \gcheck \\
        \bottomrule
    \end{tabular}
    \caption{Comparison of capabilities. General-purpose agents \cite{yang2025r, ou2025automind} support collaborative coding but lack biological priors, model exploration, and heterogeneity data handling. Specialized cell scientists \cite{tang2025cellforge} use biological priors and model exploration, but are limited to standardized data. \agentname unifies all four functions, enabling end-to-end modeling across heterogeneous datasets.}
    \label{tab:capability_comparison}
\end{table*}

\subsection{Semantic Heterogeneity Data Unifier} 
\label{sec:semantic_unification}

Semantic heterogeneity hinders scalable virtual cell modeling. While datasets share the same \texttt{h5ad} format, they exhibit divergent metadata schemas—ranging from inconsistent perturbation syntax (e.g., ``CRISPRi-KRAS'' vs. ``KRAS knockdown''), non-standardized gene identifiers (e.g., ``DDX11L16'' vs. ``ENSG00000292371''), to inconsistent variable names (e.g., ``CellType'' vs. ``cell\_line''). These misalignments blocks direct model transfer and necessitates costly manual reconciliation. 

To automate the whole pipeline, we introduce the \textbf{Semantic Heterogeneity Solver}. Instead of relying on rigid rules, this module prompts a frozen LLM with raw field descriptors to infer a canonical JSON mapping specification $\mathcal{M}$. This mapping captures both direct field aliasing and dynamic logic expressions (e.g., extracting dose values from composite strings). By executing this specification, the solver projects disparate raw datasets $\mathcal{D}_{\text{raw}}$ into a strictly unified interface $\mathcal{D}_{\text{unified}}$ without manual intervention. The formal definition of this alignment algorithm and prompt details are provided in Appendix~\ref{app:semantic_algo}.

\subsection{Statistical Heterogeneity Solver: Adaptive MCTS with Hierarchical Action Space}
\label{sec:mcts}

To bridge the gap between known biology and novel perturbations, \agentname employs an adaptive Monte Carlo Tree Search (MCTS) engine. Rather than treating code generation as a flat sequence prediction task, we frame it as a structured search for the optimal statistical inductive bias.

\paragraph{Initialization via Historical Priors.}
To accelerate exploration, \agentname leverages a meta-initialization strategy. The agent firstly maps the current task to scientifically analogous historical experiments stored in a knowledge base (see Appendix~\ref{app:retrieval_details} for details). Based on the semantic similarity to these priors, we calculate a retrieval confidence score $\rho = \max_k s_k$.
If reliable priors exist ($\rho > \tau$, \textit{In-Distribution}), the tree is \textbf{warm-started} with a retrieved architecture $\boldsymbol{\varepsilon}_0$ to focus on local refinement. Conversely, under severe distribution shift ($\rho \le \tau$), the retrieval is bypassed to prevent negative transfer, initializing the tree from a generic ``Tabula Rasa'' state for \textit{ab initio} exploration.

\paragraph{Hierarchical Action Space for Statistical Alignment.}
Navigating the vast space of executable code is computationally intractable and often fails to capture the high-level structural shifts required by diverse biological distributions. To address this, we structure the action space $\mathcal{A}$ into a three-level hierarchy (see Figure~2), explicitly designed to align model capacity with data statistical properties: 
(1) \textbf{Modeling Paradigm (Macro-Level, Strategy Space)}: This level decides the fundamental statistical assumption. For instance, the agent can choose a \textit{Generative} approach (e.g., cVAE / Flow) to model the manifold of sparse, high-dimensional gene data, or a \textit{Discriminative} approach (e.g., Regression) for continuous, dense drug-response surfaces.
(2) \textbf{Architectural Backbone (Meso-Level, Model Space)}: Selecting topologies (e.g., ResNet, GatedMLP, Transformers) that best capture the feature interactions of the specific dataset.
(3) \textbf{Optimization Refinement (Micro-Level, Engineering Space)}: Fine-tuning loss functions (e.g., Huber vs. MSE) and hyperparameters to handle outliers and noise.
This hierarchical decomposition enables the agent to ``reason'' about statistical heterogeneity from top-down—first determining the distribution type, then the structural complexity—ensuring that the synthesized models are not just syntactically valid, but statistically adaptive to the underlying biological variation.

\paragraph{Search Process.} The MCTS module executes following four phases iteratively:

\textbf{1. Selection via Optimistic UCT.}
The agent traverses the tree using a modified Upper Confidence Bound to prioritize high-potential branches:
\begin{equation}
    \text{UCT}(s') = Q_{\text{mix}}(s') + C \cdot \sqrt{\frac{\ln N(s)}{N(s') + \epsilon}},
\end{equation}
where $Q_{\text{mix}}$ balances the maximum and average observed rewards to encourage exploration of strategies with high peak performance despite initial instability.

\textbf{2. Expansion.}
Upon reaching a leaf, the LLM instantiates specific code templates. Actions are biased towards refining the retrieved structure in Warm-Start mode, or exploring the full spectrum of strategies in \textit{Ab Initio} mode.

\textbf{3. High-Fidelity Simulation.}
We execute a full training protocol to evaluate a multi-objective reward:
\begin{equation}
    R(s) = w_p \cdot \mathcal{M}_{\text{val}}(s) + w_e \cdot f_{\text{time}}(T_{\text{exec}}).
\end{equation}
The validation term $\mathcal{M}_{\text{val}}$ captures the normalized DeltaPCC (set to 0 if execution fails). Crucially, the efficiency term $f_{\text{time}}$ penalizes execution times exceeding a baseline using a piecewise linear decay function (detailed definition in Appendix), ensuring solutions are both accurate and computationally efficient.

\textbf{4. Backpropagation.}
The statistics ($N(s), \bar{Q}, Q_{\max}$) are propagated to the root using standard MCTS updates \cite{browne2012survey}, dynamically refining the agent's belief about the optimal architectural path. Note that the hierarchical structure imposes a directed search: once a higher-level decision (Paradigm and Model Structure) is fixed for a branch, the agent proceeds to lower-level refinements (Hyperparameter Refinement or Loss Funciton Refinement) without reverting the parent decision within that specific trajectory, ensuring search efficiency.


\paragraph{Datasets.} 
To benchmark generalization under diverse \textbf{statistical heterogeneities}, we utilize four datasets from three resources representing distinct distributional shifts. We employ the Srivatsan ~\cite{srivatsan2020massively} dataset to simulate \textit{continuous covariate shifts} driven by chemical perturbations across varying dosages. Conversely, we use Adamson~\cite{adamson2016multiplexed} and Norman~\cite{norman2019exploring} datasets to introduce \textit{discrete structural shifts} arising from high-dimensional CRISPR gene edits. For Srivatsan dataset, we construct splits for both unseen cell and unseen perturbation tasks to test OOD robustness. Regarding Norman, we utilize the single-perturbation subset to isolate the effects of individual gene knockouts, ensuring a controlled evaluation of structural generalization consistency. 

\paragraph{Baselines.} We compare against state-of-the-art specialized models that are widely adopted for perturbation prediction, including Biolord~\cite{piran2024disentanglement}, Sams VAE~\cite{bereket2023modelling} and CPA~\cite{lotfollahi2021compositional} for aforementioned gene and drug perturbation prediction tasks.

\begin{table*}[t]
        \small 
        \centering
        \setlength{\tabcolsep}{8pt} 
        
        \begin{tabular}{l ccc c}
            \toprule
            \multirow{2}{*}{\textbf{Agents}} & \multicolumn{3}{c}{\textbf{Error Type}} & \multirow{2}{*}{\textbf{Valid Execution Rate}} \\
            \cmidrule(lr){2-4} 
            
             & Preprocess Error & Model Error & Hallucinated Success & \\
            \midrule
    
            AIDE~\cite{jiang2025aide}        & 35\% & 50\% & 15\% & 0\% \\
            R\&D Agent~\cite{yang2025r}  & 45\% & 30\% & 25\% & 0\% \\

            \textbf{\agentname} & \textbf{0\%} & \textbf{5\%} & \textbf{0\%} & \textbf{95\%} \\
    
            \bottomrule
        \end{tabular}
    \caption{
        \textbf{Superiority in Handling Semantic Heterogeneity.} While general coding agents (AIDE, R\&D Agent) failed all 20 virtual cell modeling trials despite detailed manual guidance, \agentname achieved a 95\% success rate. This underscores its robust capability in resolving data heterogeneity and automating preprocessing. 
    }
    \label{tab:error_type_comparison}
\end{table*}

\begin{figure*}[t]
    \begin{center}
    \centerline{\includegraphics[width=0.89\linewidth]{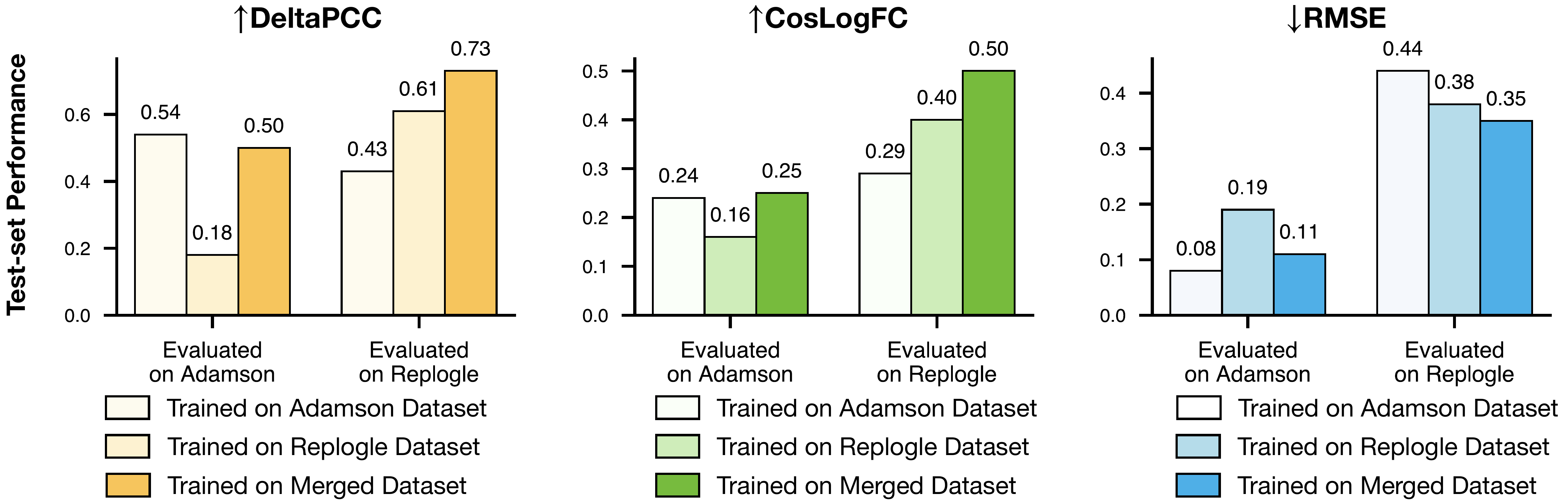}}
    \caption{\textbf{\agentname successfully handles semantic heterogeneity, and enables synergistic dataset scaling.} We compared the generalization performance of models trained on single-source Adamson \cite{adamson2016multiplexed} and Replogle \cite{replogle2022mapping} datasets against the joint dataset harmonized by \agentname's Semantic Unifier.  Models were evaluated on held-out test sets under a unified separate cross-validation protocol.}
    \label{fig:scaling_performance}
    \end{center}
    \vskip -0.3in
\end{figure*}

\section{Experiments}
\label{sec:experiments}

We evaluate HarmonyCell on a comprehensive suite of single-cell perturbation tasks, systematically verifying its capability to handle diverse biological mechanisms and distinct data heterogeneity. Through extensive experiments, we aim to answer the following questions:

\textbf{RQ1: Semantic Heterogeneity Resilience.} Can HarmonyCell effectively resolve semantic conflicts in uncurated metadata to ensure reliable end-to-end model execution compared to general coding agents?

\textbf{RQ2: Synergistic Data Scaling.} Does automated semantic unification enable predictive gains by synergistically integrating heterogeneous datasets from disparate sources?

\textbf{RQ3: Statistical Generalization Efficiency.} How does HarmonyCell perform across diverse biological distribution shifts (e.g., unseen perturbations/cells) compared to state-of-the-art expert-designed models?
\subsection{Semantic Heterogeneity Solver}

We evaluated \agentname against two general coding agents: AIDE~\cite{jiang2025aide} and R\&D Agent~\cite{yang2025r}. CellForge~\cite{tang2025cellforge} was not considered in our experiments because we were unable to reproduce the released version of its open-source repository. As shown in Table~\ref{tab:error_type_comparison}, quantitative results for CellForge are absent (``--'') as its open-source repository were irreproducible, preventing valid experimental evaluation. Consequently, we focused on two general agents, which universally failed to handle heterogeneous data through 20 trials, yielding a \textbf{0\% valid execution rate}. Though given manually mapped data preprocessing instructions and model insights, these agents struggled primarily with data ingestion, exhibiting high preprocess error rates (35\%--45\%) due to schema misalignments. Alarmingly, they also showed significant hallucinated success rates (15\%--25\%), falsely manupulate codebase to either create random data or non-compliant validation set calculation methods to ensure task completion.

In contrast, \agentname achieved a \textbf{95\% valid execution rate} with \textbf{0\% preprocessing errors} given no manual mapping and model insights. By employing the Semantic Unifier to induce canonical mappings prior to code generation, \agentname effectively eliminates semantic heterogeneities. This result demonstrates that while general agents fail "in the wild," \agentname provides the necessary semantic resilience for reliable virtual cell modeling.

\subsection{From Heterogeneity to Scalability: Automated Data Unification}
\label{sec:semantic_resilience}

We designed a Data Integration Challenge combining the Adamson~\cite{adamson2016multiplexed} and Replogle~\cite{replogle2022mapping} datasets. As both datasets involve CRISPRi perturbations in K562 cells, they share overlapping biological contexts yet differ in metadata schemas, making them ideal candidates for evaluating automated alignment.

We quantify whether the agent-unified data leads to predictive gains under a \textit{Separate Evaluation Protocol}. Specifically, we trained models on the joint dataset $\mathcal{D}_{union}$ (automatically aligned by \agentname) and evaluated them strictly on the held-out test sets of each source individually.

As illustrated in Figure~\ref{fig:scaling_performance}, \agentname successfully demonstrates the value of data scaling. Compared to baselines trained on single-source data, the model trained on the agent-harmonized dataset shows consistent performance improvements. Notably, we observe significant \textbf{positive transfer}, where the inclusion of Replogle data enhances the model's generalization on Adamson's unseen perturbations (outperforming the in-domain specialist). This result confirms that \agentname does not just resolve syntactic conflicts but effectively harmonizes underlying biological signals, streamlining the path to robust data scaling without manual intervention.

\begin{table*}[htbp]
    \small
    \centering
    \setlength{\tabcolsep}{4pt} 
    
    \begin{tabular}{l ccc ccc}
        \toprule
        
        & \textbf{CosLogFC} $\uparrow$ & \textbf{RMSE} $\downarrow$ & \textbf{DeltaPCC} $\uparrow$ 
        & \textbf{CosLogFC} $\uparrow$ & \textbf{RMSE} $\downarrow$ & \textbf{DeltaPCC} $\uparrow$ \\
        \midrule

        \rowcolor[rgb]{0.98, 0.94, 0.82} 
        \textbf{\textit{Gene Perturbation}} & 
        \multicolumn{3}{c}{$\circ$ \textit{Norman \cite{norman2019exploring}}} & 
        \multicolumn{3}{c}{$\circ$ \textit{Adamson \cite{adamson2016multiplexed}}} \\
        \addlinespace[0.3em] 
        
        Biolord \cite{piran2024disentanglement}     & 0.57 & \underline{0.05} & 0.42 & \underline{0.52} & \underline{0.02} & 0.54 \\
        Sams VAE \cite{bereket2023modelling}          & 0.58 & \underline{0.05} & 0.44 & 0.51 & \underline{0.02} & \underline{0.55} \\
        CPA \cite{lotfollahi2021compositional}      & 0.55 & \underline{0.05} & 0.39 & 0.41 & 0.03 & 0.32 \\
        HarmonyCell                                 & \textbf{\underline{0.61}} & \textbf{0.09} & \textbf{\underline{0.62}} & \textbf{0.32} & \textbf{0.09} & \textbf{0.49} \\

        \addlinespace[0.8em] 
        
        \rowcolor[rgb]{0.85, 0.93, 0.96} 
        \textbf{\textit{Drug Perturbation}} & 
        \multicolumn{3}{c}{$\circ$ \textit{Srivatsan-Sciplex2 \cite{srivatsan2020massively}}} & 
        \multicolumn{3}{c}{$\star$ \textit{Srivatsan-Sciplex3 \cite{srivatsan2020massively}}} \\
        \addlinespace[0.3em] 
        
        Biolord \cite{piran2024disentanglement}     & \underline{0.14} & \underline{0.05} & 0.20 & 0.08 & 0.14 & 0.06 \\
        Sams VAE \cite{bereket2023modelling}          & 0.12 & \underline{0.05} & 0.19 & 0.07 & 0.18 & 0.04 \\
        CPA \cite{lotfollahi2021compositional}      & 0.12 & \underline{0.05} & 0.16 & 0.08 & 0.15 & 0.05 \\
        HarmonyCell                                 & \textbf{0.06} & \textbf{0.09} & \textbf{\underline{0.26}} & \textbf{\underline{0.10}} & \textbf{\underline{0.07}} & \textbf{\underline{0.29}} \\

        \midrule
        \multicolumn{7}{c}{\textbf{Task Category:}\quad \footnotesize $\circ$: Unseen Perturbation Prediction \qquad $\star$: Unseen Cell Prediction} \\
        \bottomrule
    \end{tabular}
    \caption{
    \textbf{Performance on disjoint datasets under statistical heterogeneity.} 
    The model is evaluated on datasets with distinct statistical heterogeneities (Unseen Perturbation and Unseen Cell). \agentname consistently matches or exceeds expert-level baseline performance, particularly in maintaining correlation (DeltaPCC) across distribution shifts. Underlined results indicate the best performance among baselines, while bold results indicate \agentname's performance. The uparrow $\uparrow$ stands for higher value, while the downarrow $\downarrow$ stands for lower value for better performance of a metric.
    }
    \label{tab:performance}
\end{table*}

As illustrated in Figure~\ref{fig:scaling_performance}, single-source baselines exhibit limited cross-domain generalization (the Replogle-trained model suffers a significant performance drop when evaluated on Adamson). In contrast, the model trained on the agent-harmonized dataset demonstrates robust learnability across domains. It effectively bridges the generalization gap on Adamson tasks while achieving a synergistic effect on Replogle tasks, where the unified model outperforms even the in-domain specialist (DeltaPCC 0.73 vs. 0.61). This result confirms that \agentname\ effectively harmonizes underlying biological signals, streamlining the path to robust data scaling across heterogeneous laboratories.

\subsection{Generalization under Statistical Heterogeneity}
\label{sec:statistical_generalization}

\begin{figure}[ht] 
    \centering 
    \includegraphics[trim=0 10pt 0 10pt, clip, width=0.98\columnwidth]{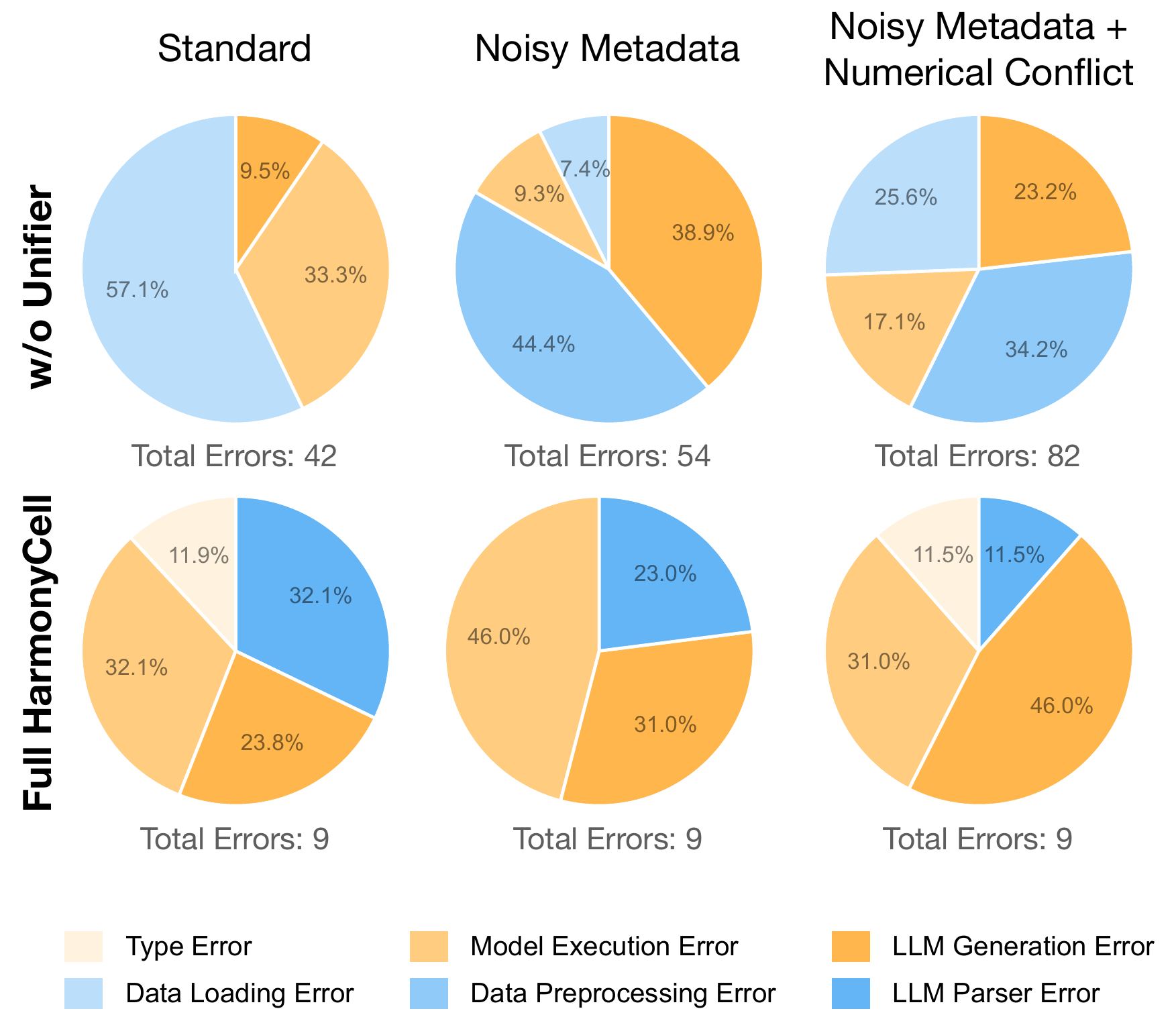}
    
    \vspace{-5pt} 
    \caption{\textbf{Ablation Study: Necessity of Semantic Unifier.} During execution, HarmonyCell containing the Semantic Unifier exhibits a more stable and error-free execution workflow compared to agents without the module.}
    \label{fig:piechart}
    
    \vspace{5pt} 
\end{figure}

Beyond semantic unification, the fundamental impediment to scalable modeling lies in adapting to distinct biological distributions, formally defined as \textbf{Statistical Heterogeneity}. Table~\ref{tab:performance} demonstrates that \agentname successfully bridges the gap between diverse modeling domains for virtual cell tasks. We interpret these quantitative gains not merely as numerical improvements, but as empirical evidence of the MCTS engine's capability to autonomously synthesize architectures tailored to specific distributional challenges:

\textbf{Adaptation to Continuous Covariate Shift (Drug Perturbation).}  The Srivatsan~\cite{srivatsan2020massively} benchmarks epitomize statistical heterogeneity arising from continuous, dose-dependent covariate shifts. As detailed in Table~\ref{tab:performance}, \agentname consistently matches or surpasses specialized baselines (e.g., CPA, Biolord) across both unseen perturbation and unseen cell regimes. Most notably, on the Srivatsan-Sciplex3 dataset, \agentname attains a superior correlation coefficient (DeltaPCC: \textbf{0.29}) while maintaining minimal reconstruction error (RMSE: \textbf{0.07}). This performance substantiates the agent's capacity to accurately model non-linear dose-response manifolds without manual architecture selection, effectively resolving continuous distributional shifts.

\textbf{Adaptation to Discrete Combinatorial Shift (Gene Perturbation).}  In contrast, the Norman, and Adamson datasets present ``structural'' statistical heterogeneity, characterized by high-dimensional sparsity and discrete gene interactions. Under these rigorous Out-of-Distribution (OOD) conditions, \agentname exhibits robust generalization capabilities. Specifically, on the Norman~\cite{norman2019exploring} dataset, \agentname achieves a CosLogFC of 0.61 and DeltaPCC of 0.62, significantly outperforming the leading baseline (CosLogFC of 0.58 and DeltaPCC of 0.44). This result underscores the agent's ability to capture intricate genetic dependency patterns and dynamically adapt its statistical inductive bias to the underlying data structure, delivering state-of-the-art accuracy through automated adaptation.

\subsection{Ablation Study}
\label{sec:ablation}

To verify the necessity of our architectural components, we conduct controlled experiments dissecting the \textbf{Semantic Unifier} and the \textbf{Hierarchical Action Space}.

\paragraph{Impact of Semantic Unifier on Data Resilience.}

We constructed three variants of the Norman~\cite{norman2019exploring} dataset with increasing schema noise: (1) \textit{Standard} (clean, log1p normalized values); (2) \textit{Noisy Metadata} (aliased control fields); and (3) \textit{Noisy Data + Numerical Conflict} (unnormalized values). We compared \agentname against a baseline without the Semantic Unifier under a standard Hierarchical MCTS described in Section~\ref{sec:mcts}. As quantified in Figure~\ref{fig:piechart}, the ablated agent was dominated by multiple running errors which mainly appear in data processing and model design. In contrast, the full \agentname successfully eliminated these ingestion failures, restricting errors solely to the downstream modeling phase. This confirms that the Semantic Unifier is not merely an optimization, but a prerequisite for resilience against semantic heterogeneity.

\paragraph{Necessity of Hierarchical Search for Statistical Alignment.}

We investigated whether a hierarchical action space is strictly necessary compared to a simpler MCTS baseline, where atomic operations ``Refine Hyperparameters'' and ``Refine Loss'' are searched after determination of the learning paradigm. All ablation experiments are conducted on the Norman Dataset with \textbf{Gene Perturbation (OOD)} task for 16 node expansions. Since CosLogFC and DeltaPCC are better indicators of the sensitivity of virtual cell models to changes in distribution levels than RMSE, we used these two indicators as the benchmarks for comparing our final experiments with the baseline.

\begin{figure}[t]
    \begin{center}
    \centerline{\includegraphics[width=1.08\columnwidth]{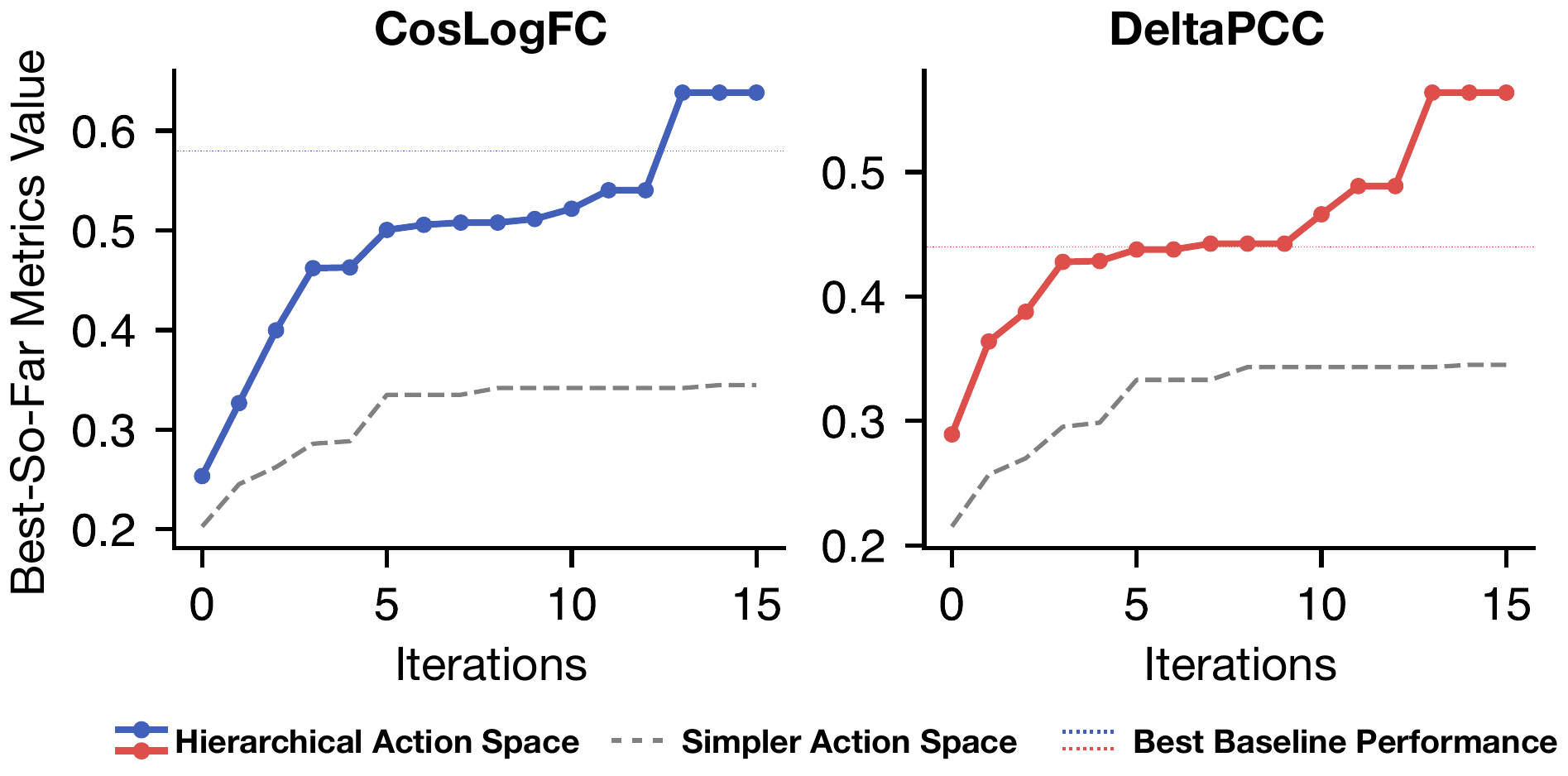}}
    \caption{\textbf{Ablation Study: Necessity of Hierarchical Action Space.} \agentname achieves superior convergence speed and accuracy compared to the non-hierarchical ablated agents, effectively surpasses the state-of-the-art specialized baseline models.}
    \label{fig:mcts_ablation}
    \label{fig:ablation_mcts}
    \end{center}
    \vskip -0.35in
    \end{figure}

As shown in Figure~\ref{fig:ablation_mcts}, the agent without hierarchical search space became trapped in local optima without improving generalization ability. Moreover, the performance improvement is significantly lower than that of \agentname using the Hierarchical Search Space. In contrast, \agentname's hierarchical action space enforced a top-down decision process, enabling an faster convergence rate and more robust performance in both metrics, surpasses the state-of-the-art specialized baseline models by over 10\% in CosLogFC and 20\% in DeltaPCC.

\subsection{Case Study: Evolutionary Architecture Design}
\label{sec:case_study}

To explicitly visualize the decision-making process of \agentname, we analyze a standard MCTS search trajectory (with 16 node expansions) on Norman dataset~\cite{norman2019exploring} with 32 nodes and 5 hours' time limit.

\begin{figure}[t]
    \begin{center}
    \centerline{\includegraphics[width=0.99\columnwidth]{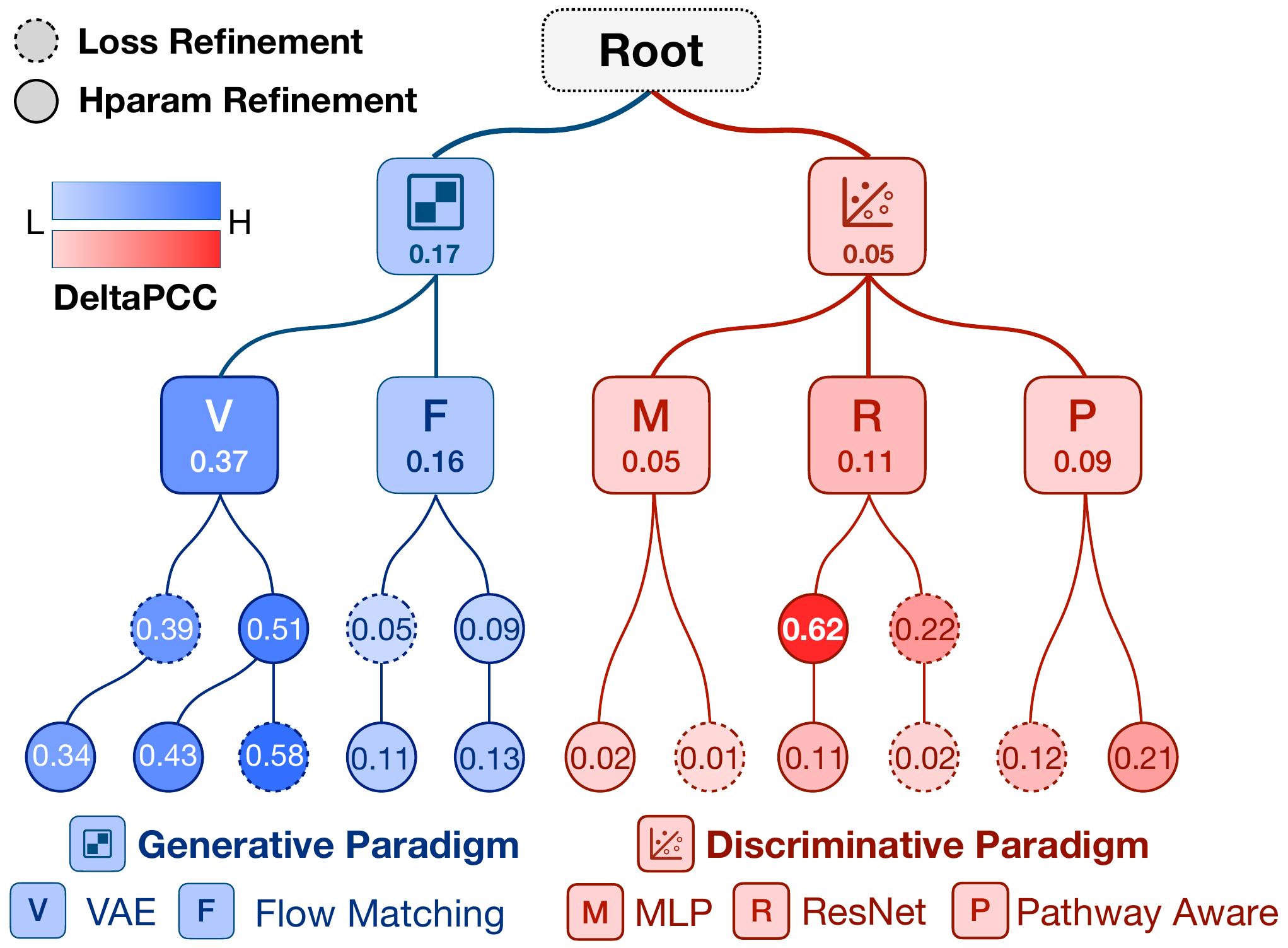}}
    \caption{\textbf{Case Study: MCTS Exploration for Norman dataset.} The number inside each node represents the DeltaPCC validation value of the model at that node.}
    \label{fig:casestudy}
    \end{center}
    \vskip -0.5in
    \end{figure}

\textbf{Decision Path Analysis.} 
The search tree illustrates how \agentname navigates the hierarchical action space to find the global optimum despite misleading initial signals. In all 24 iterations, the model finally found the global optimum at the 8th iteration whose DeltaPCC validation value is 0.62, surpassing our best-performing baseline (Sams VAE, 0.44). The detailed trajectory is shown in Figure~\ref{fig:casestudy}:

\textbf{1. Initial Generative Preference:} At the root level, the agent initially favored the \textit{Generative Paradigm} (left branch) over the \textit{Discriminative Paradigm}, likely due to the latter's training instability on sparse data. It extensively explored the \textit{VAE (V)} backbone, achieving a high performance of 0.58 through \textit{Hyperparameter Refinement}.
    
\textbf{2. Exploration of Discriminative Subspace:} Despite the lower average reward in the Discriminative branch, the MCTS exploration mechanism (driven by the UCT algorithm) continued to probe this subspace and finally find a node with relatively high performance (DeltaPCC = 0.58), which already reach the expert-level performance when compared to specialized baseline models.
    
\textbf{3. Critical Leap-Out from Local Optima:} Although generative models can achieve optimal metrics for most nodes, due to the exploratory-exploitation strategy of MCTS, the agentname still shifts to an exploratory discriminative model. It identified the \textit{ResNet (R)} backbone as a potential candidate where the node with the best performance was found eventually. Within the ResNet branch, the agent achieved the global maximum score (DeltaPCC = 0.62) after a hyperparameter refinement step. 

This trajectory highlights a key advantage of \textit{hierarchical search space} design in MCTS engine of \agentname. Unlike greedy search algorithms, it exploits the stable performance of generative models while avoiding local optima by successfully identifying specific, high-performing discriminative architectures. Enable expert-level model performance for given virtual cell tasks without human intervention.

\section{Conclusion}
This paper introduces \agentname, an framework designed to resolve the \textbf{dual heterogeneity} bottleneck in automated virtual cell modeling: semantic schema inconsistencies and statistical distribution shifts. Decoupling the workflow into an LLM-driven optimization, our approach autonomously converts uncurated metadata into robust architectures. Extensive benchmarks demonstrate that \agentname achieves expert-level accuracy and a 95\% valid execution rate across semantically disparate datasets. The results validate our dual-track orchestration as a scalable foundation for automated scientific discovery in the ``virtual cell'' era.

\section*{Impact Statement}
This paper presents work whose goal is to advance the field of Machine Learning. There are many potential societal consequences  of our work, none which we feel must be specifically highlighted here.

\nocite{langley00}

\bibliography{example_paper}
\bibliographystyle{icml2025}

\newpage
\appendix
\onecolumn
\section{Implementation Details of HarmonyCell}
\label{app:implementation}

In this section, we provide the specific mathematical formulations and hyperparameter settings used in the HarmonyCell framework, specifically regarding the MCTS engine and the hierarchical action space.

\subsection{Metrics Calculation Protocols} \label{supp:metrics}
To strictly evaluate the model's performance in predicting perturbation responses, we conduct evaluations at the pseudo-bulk level. Let $y_p \in \mathbb{R}^G$ and $\hat{y}_p \in \mathbb{R}^G$ denote the ground-truth and predicted mean gene expression vectors for a specific perturbation $p$, averaged across all cells in that condition. Similarly, let $y_{ctrl} \in \mathbb{R}^G$ represent the mean expression vector of the control (unperturbed) cells. The dimensionality $G$ corresponds to the number of highly variable genes (HVGs).

We define the true perturbation effect (shift vector) as $\delta_p = y_p - y_{ctrl}$ and the predicted effect as $\hat{\delta}_p = \hat{y}_p - y_{ctrl}$. The calculation protocols for the three metrics are defined as follows:

\paragraph{1. Root Mean Square Error (RMSE)}
RMSE measures the global reconstruction error between the predicted and true pseudo-bulk expression profiles. It is calculated as:
\begin{equation}
    \text{RMSE}(y_p, \hat{y}_p) = \sqrt{\frac{1}{G} \sum_{j=1}^{G} (y_{p,j} - \hat{y}_{p,j})^2}
\end{equation}
where $y_{p,j}$ and $\hat{y}_{p,j}$ are the expression values of the $j$-th gene. Lower RMSE indicates better predictive accuracy in the absolute gene expression space.

\paragraph{2. Pearson Correlation of Delta (DeltaPCC)}
DeltaPCC assesses the linearity and correlation of the predicted gene expression changes relative to the ground truth changes. It is computed as the Pearson correlation coefficient between the true shift vector $\delta_p$ and the predicted shift vector $\hat{\delta}_p$:
\begin{equation}
    \text{DeltaPCC}(\delta_p, \hat{\delta}_p) = \frac{\text{Cov}(\delta_p, \hat{\delta}_p)}{\sigma_{\delta_p} \sigma_{\hat{\delta}_p}} = \frac{\sum_{j=1}^{G} (\delta_{p,j} - \bar{\delta}_p)(\hat{\delta}_{p,j} - \bar{\hat{\delta}}_p)}{\sqrt{\sum_{j=1}^{G} (\delta_{p,j} - \bar{\delta}_p)^2} \sqrt{\sum_{j=1}^{G} (\hat{\delta}_{p,j} - \bar{\hat{\delta}}_p)^2}}
\end{equation}
where $\bar{\delta}_p$ and $\bar{\hat{\delta}}_p$ are the means of the elements in vectors $\delta_p$ and $\hat{\delta}_p$, respectively. A higher DeltaPCC indicates that the model correctly captures the pattern of gene regulation changes.

\paragraph{3. Cosine Similarity of Log Fold Change (CosLogFC)}
CosLogFC evaluates the directional fidelity of the predicted perturbation effects in the high-dimensional gene space. Since the data is log-normalized, the shift vector $\delta_p$ effectively represents the Log Fold Change. The metric is defined as:
\begin{equation}
    \text{CosLogFC}(\delta_p, \hat{\delta}_p) = \frac{\delta_p \cdot \hat{\delta}_p}{\|\delta_p\|_2 \|\hat{\delta}_p\|_2} = \frac{\sum_{j=1}^{G} \delta_{p,j} \hat{\delta}_{p,j}}{\sqrt{\sum_{j=1}^{G} \delta_{p,j}^2} \sqrt{\sum_{j=1}^{G} \hat{\delta}_{p,j}^2}}
\end{equation}
Values closer to 1 indicate that the predicted perturbation vector aligns perfectly with the direction of the true biological change.

\subsection{Computational Efficiency Reward Formulation}
\label{app:time_reward}

To balance model performance with computational cost, we introduced a time-efficiency reward component $f_{\text{time}}(T_{\text{exec}})$ in Eq.~(9) of the main text. This function is designed to strictly penalize models that exceed the execution time of a standard baseline ($T_{\text{root}}$).

The function $f_{\text{time}}(T_{\text{exec}})$ is defined as a piecewise linear decay function relative to the normalized time ratio $t = T_{\text{exec}} / T_{\text{root}}$:

\begin{equation}
    f_{\text{time}}(t) = \begin{cases} 
        1.0 & \text{if } t \leq 0.8 \\[8pt]  
        1.0 - 0.2 \cdot \dfrac{t - 0.8}{0.2} & \text{if } 0.8 < t \leq 1.0 \\[10pt] 
        0.8 - 0.3 \cdot \dfrac{t - 1.0}{0.5} & \text{if } 1.0 < t \leq 1.5 \\[10pt]
        \max\left(0, 0.5 - 0.5 \cdot \dfrac{t - 1.5}{1.5}\right) & \text{if } t > 1.5
    \end{cases}
\end{equation}

This design ensures that:
\begin{itemize}
    \item Models significantly faster than the baseline ($t \le 0.8$) receive the maximum efficiency reward ($1.0$).
    \item Models slightly slower than the optimal range ($0.8 < t \le 1.0$) face a moderate penalty.
    \item Models exceeding the baseline time by 50\% ($t > 1.5$) are heavily penalized, discouraging computationally expensive architectures.
\end{itemize}

\subsection{Detailed Hierarchical Action Space}
\label{app:action_space}

The MCTS engine explores a structured, three-level hierarchical action space to synthesize virtual cell models. The specific candidate operations available at each level are listed below:

\begin{enumerate}
    \item \textbf{Level 1: Modeling Paradigm (Strategy)}
    \begin{itemize}
        \item \textit{Discriminative Strategy:} Deterministic mapping $f(x_{\text{ctrl}}, p) \to \Delta x$, focusing on high-fidelity point-to-point prediction of state changes. Best for maximizing Delta-PCC with dense, high-quality data.
        \item \textit{Generative Strategy:} Probabilistic modeling of the distribution $P(x|p)$, capturing underlying distributions and continuous velocity fields of gene expression. Selected for sparse or noisy data regimes.
    \end{itemize}
    
    \item \textbf{Level 2: Architectural Backbone (Structure)}
    \begin{itemize}
        \item \textit{For Discriminative Strategy:}
        \begin{itemize}
            \item \textit{ResNet Block:} Deep residual networks for capturing non-linear gene interactions with improved gradient flow.
            \item \textit{Gated MLP:} Sparse architectures with gating mechanisms, efficient for high-dimensional gene expression.
            \item \textit{Pathway-Aware Masked:} Incorporates biological priors through pathway-aware layers, enabling gene alignment based on known biological pathways.
        \end{itemize}
        \item \textit{For Generative Strategy:}
        \begin{itemize}
            \item \textit{Conditional VAE (cVAE):} Variational autoencoder architecture for generative modeling of perturbation effects.
            \item \textit{Flow Matching:} Continuous normalizing flows for modeling complex perturbation trajectories and velocity fields.
        \end{itemize}
    \end{itemize}
    
    \item \textbf{Level 3: Optimization Refinement (Engineering)}
    \begin{itemize}
        \item \textit{Hyperparameter Refinement:} Tuning learning rate, weight decay, dropout, and other hyperparameters to optimize training dynamics.
        \item \textit{Loss Function Refinement:} Suggesting the use of Huber loss instead of standard MSE loss to increase robustness against outliers in sparse gene expression data.
    \end{itemize}
    
    \item \textbf{Special Action: Debug} - Available at any depth when a node has execution errors
    \begin{itemize}
        \item \textit{Debug:} Automatic error fixing through LLM-based debugging, allowing the agent to recover from code generation or execution failures.
    \end{itemize}
\end{enumerate}

The hierarchical structure ensures that strategy selection precedes architecture design, which in turn precedes refinement operations. This constraint prevents invalid action sequences and guides the search toward semantically meaningful model configurations.

\subsection{Hyperparameter Settings}
\label{app:hyperparams}

We list the key hyperparameters used in the search process and reward calculation in Table~\ref{tab:hyperparams}.

\begin{table}[h]
\centering
\caption{Hyperparameters for HarmonyCell MCTS Engine.}
\label{tab:hyperparams}
\begin{tabular}{lccc}
\toprule
\textbf{Component} & \textbf{Symbol} & \textbf{Value} & \textbf{Description} \\
\midrule
\multirow{2}{*}{Retrieval (RAG)} & $\tau_{filter}$ & 0.3 & Threshold for filtering irrelevant historical tasks \\
& $m$ & 3 & Number of top neighbors retrieved \\
\midrule
\multirow{3}{*}{MCTS Selection} & $\alpha$ & 0.7 & Weight for Optimistic Value ($Q_{mix} = \alpha Q_{max} + (1-\alpha)\bar{Q}$) \\
& $C$ & 1.0 & Exploration constant in UCT formula \\
& $N_{sim}$ & 32 & Number of MCTS simulation iterations \\
\midrule
\multirow{2}{*}{Reward Function} & $w_p$ & 0.8 & Weight for validation performance ($\Delta$PCC) \\
& $w_e$ & 0.2 & Weight for computational efficiency ($f_{time}$) \\
\bottomrule
\end{tabular}
\end{table}

It is noted that the time limit is set to 5 hours in all experiments, which is sufficient for the MCTS engine to find the global optimum. If the time limit is exceeded, the MCTS engine will return the best model found so far and terminate the search process.

\subsection{Dataset Processing and Split Strategy}
\label{app:dataset_processing}

All datasets are processed using a unified preprocessing pipeline, including log2-transformation, normalization, and selection of the top 2,000 highly variable genes (HVGs) within each dataset.
For drug-perturbation tasks, we use two datasets from Srivatsan \cite{srivatsan2020massively}. Because most baseline models do not natively model dose–response, we fix all experiments to the 10 $\mu$M dose.

\begin{itemize}
    \item Sci-Plex2 contains roughly 7k A549 cell lines across four drug perturbations. We randomly assign three drugs to the training set, with the remaining drug split into validation and test, forming an drug-based unseen-perturbation task.
    \item Sci-Plex3 comprises approximately 1.6 $\times$ 10$^5$ cells from A549, K562, and MCF7 across 188 drugs.  We subset metadata to the 24-hour time point (dose fixed at 10 $\mu$M). K562 and MCF7 profiles are used for training, while A549 is reserved for validation and test, establishing an unseen-cell setting.
\end{itemize}

For gene-perturbation tasks, we use three datasets to establish unseen-perturbation tasks.
\begin{itemize}
    \item Adamson2016 (K562) includes around 6 $\times$ 10$^3$ CRISPRi knockdown cells, with 5 perturbations used for training and 2 for validation/test.
    \item Norman2019 (K562) provides roughly 7 $\times$ 10$^4$ cells with 105 single-gene CRISPRi perturbations, split 8:2 at the perturbation level.
    \item Replogle2022 (K562 essential) contains over 1.1 $\times$ 10$^5$ CRISPRi perturbation profiles. We select the 300 most abundant perturbations and apply the same 8:2 split. This dataset is additionally used to evaluate model performance under cross-dataset merging, where it is combined with other gene-perturbation datasets to test robustness against heterogeneous data integration.
\end{itemize}
Across all unseen-perturbation settings, control cells are partitioned using identical train/validation/test ratios as perturbed cells. All datasets originate from the standardized scPerturb resource~\cite{peidli2024scperturb}.

\subsection{API Usage}

To balance among time, cost and performance, we finally choose Google Gemini 3 Flash (thinking) as our LLM API backend.

\section{Semantic Unification Algorithm Details}
\label{app:semantic_algo}

In this section, we provide the formal definition of the Semantic Unification process described in Section~\ref{sec:semantic_unification}, along with an illustrative example of the LLM-generated mapping specification.

\subsection{Formal Algorithm}

The Semantic Heterogeneity Solver operates by transforming raw metadata schemas into a target canonical schema through an intermediate mapping object. The procedure is formally described in Algorithm~\ref{alg:formal_mapping_app}.

\begin{algorithm}[H] 
   \caption{Formal Semantic Unification via LLM-Guided Schema Alignment}
   \label{alg:formal_mapping_app}
\begin{algorithmic}[1]
   \STATE {\bfseries Input:} Raw dataset $\mathcal{D}_{\text{raw}} = (\mathbf{X}, \mathbf{M}_{\text{obs}}, \mathbf{M}_{\text{var}}, \mathbf{M}_{\text{obsm}})$, where $\mathbf{X}$: expression matrix; $\mathbf{M}_{\text{obs}}$: cell metadata.
   \STATE {\bfseries Target Schema:} $\mathcal{S}^{\star} = (\mathcal{S}_{\text{obs}}^{\star}, \mathcal{S}_{\text{var}}^{\star})$
   
   \STATE \textit{// Step 1: Schema Inference via LLM}
   \STATE Extract raw schema keys $\mathcal{S}^{\text{raw}} = \text{Keys}(\mathbf{M}_{\text{obs}})$.
   \STATE Construct prompt $\mathcal{P} = \Pi(\mathcal{S}^{\text{raw}}, \mathcal{S}^{\star})$ containing field descriptions and few-shot examples.
   \STATE Query frozen LLM to obtain JSON mapping:
   \[
       \mathcal{M} = \text{LLM}(\mathcal{P}) \in \mathbb{M}
   \]
   
    \STATE \textit{// Step 2: Apply Semantic Transformation}
    \FOR{each canonical key $k$ in Target Schema $\mathcal{S}^{\star}$}
        \IF{$\mathcal{M}[k]$ is \texttt{null}}
            \STATE $\widetilde{\mathbf{M}}_{\text{obs}}[k] \leftarrow \text{DefaultValue}(k)$
        \ELSIF{$\texttt{type}(\mathcal{M}[k]) == \texttt{logic}$}
            \STATE \textit{// Execute Python expression derived from LLM}
            \STATE $\widetilde{\mathbf{M}}_{\text{obs}}[k] \leftarrow \textsc{Exec}(\mathcal{M}[k].\text{expression}, \mathbf{M}_{\text{obs}})$
        \ELSE
            \STATE \textit{// Direct mapping}
            \STATE $\widetilde{\mathbf{M}}_{\text{obs}}[k] \leftarrow \mathbf{M}_{\text{obs}}[\mathcal{M}[k]]$
        \ENDIF
    \ENDFOR

   \STATE {\bfseries Output:} Unified dataset $\mathcal{D}_{\text{unified}}$ aligned to $\mathcal{S}^{\star}$.
\end{algorithmic}
\end{algorithm}

\subsection{JSON Mapping Example}

To concretely illustrate the output of the LLM for Semantic Unifier, we present a simplified example of a generated mapping object $\mathcal{M}$. This object maps a raw dataset with non-standard fields (e.g., "dose\_val", "drug\_name") to our canonical schema.

\begin{lstlisting}[language=json, basicstyle=\ttfamily\small, caption={Example JSON Mapping Specification generated by the LLM.}, label={lst:json_map}]
{
  "perturbation_type": "drug",
  "perturbation_name": {
    "type": "direct",
    "source_key": "drug_id"
  },
  "dose_value": {
    "type": "logic",
    "expression": "df['conc_um'].astype(float) * 1000", 
    "description": "Convert micromolar to nanomolar"
  },
  "cell_line": {
    "type": "direct",
    "source_key": "cell_type_annotation"
  },
  "control_status": {
    "type": "logic",
    "expression": "df['drug_id'] == 'DMSO'"
  }
}
\end{lstlisting}

\subsection{Details of Meta-Initialization Mechanism}
\label{app:retrieval_details}

In this section, we provide the mathematical formulation for the meta-initialization process mentioned in the main text.

Formally, given a new task with perturbation $p$ and covariate $c$, we generate a structured task profile $\mathbf{p}_{\text{query}}$ combined with a solution description $d_{\text{query}}$. This is embedded into a semantic space to yield $\mathbf{z}_{\text{query}} = \mathbf{E}(\mathbf{p}_{\text{query}} + d_{\text{query}})$. We maintain a knowledge base $\mathcal{K} = \{ \mathbf{e}_k \}_{k=1}^K$ recording historical task profiles, action paths, and rewards.

We retrieve the top-$m$ nearest neighbors via cosine similarity $s_k = \frac{\mathbf{z}_{\text{query}}^T \mathbf{z}_k}{\|\mathbf{z}_{\text{query}}\| \|\mathbf{z}_k\|}$. If the maximum similarity exceeds a threshold $\tau$, we leverage these priors to construct a retrieval-augmented context $\mathcal{C}_{\text{ctx}}$. Specifically, historical entries are ranked by a composite weight $w_k$ to balance relevance and performance:
\begin{equation}
    w_k = \alpha \cdot \bar{s}_k + (1-\alpha) \cdot \bar{r}_k,
\end{equation}
where $\bar{s}_k$ and $\bar{r}_k$ are the normalized semantic similarity and historical reward, respectively. This weighting ensures the context emphasizes both semantic relevance and proven historical success. The initial architecture $\boldsymbol{\varepsilon}_0$ is then synthesized by the LLM, conditioned on these weighted priors.

We also introduced a composite weight $w_k$ to rank retrieved historical tasks. This weighting mechanism ensures that the constructed context $\mathcal{C}_{\text{ctx}}$ includes examples that are not only semantically similar to the current task but also historically successful.

The specific formulation for the composite weight is:
\begin{equation}
    w_k = \alpha \cdot \underbrace{\frac{s_k - \tau_{\text{filter}}}{1 - \tau_{\text{filter}}}}_{\text{Normalized Similarity } (\bar{s}_k)} + (1-\alpha) \cdot \underbrace{\frac{r_k - r_{\min}}{r_{\max} - r_{\min}}}_{\text{Normalized Reward } (\bar{r}_k)},
\end{equation}
where:
\begin{itemize}
    \item $s_k$ is the raw cosine similarity score.
    \item $\tau_{\text{filter}} = 0.3$ is the minimum similarity threshold; entries with $s_k \le \tau_{\text{filter}}$ are discarded.
    \item $r_k$ is the recorded reward of the historical task $k$.
    \item $r_{\min}$ and $r_{\max}$ are the minimum and maximum rewards observed in the retrieved candidate set $\mathcal{N}_m$.
    \item $\alpha$ is a balancing hyperparameter, set to $0.5$ in our experiments.
\end{itemize}

Based on this weight, we select the top-$m$ entries (where $m=3$) to form the prompt context:
\begin{equation}
    \mathcal{N}_m(p, c) = \operatorname*{argtopk}_{k \in [K], s_k > \tau_{\text{filter}}} \; w_k.
\end{equation}

\section{Prompts for Key Modules in \agentname}
\label{app:prompts}

In this section, we present the specific prompts used in the Semantic Unifier and MCTS Engine modules.

\begin{promptbox}[Prompt for Dataset Previewing]
    
    \agentrole{You are a professional computational biology data expert. Please perform a "semantic profiling" of this dataset based on the following data audit information, following the USCP-DS v1.0 standard.

    **Core Objective:** You must output a structured JSON mapping dictionary to guide subsequent automated data preprocessing.
    
    \{uscp\_spec\}
    
    \{field\_semantics\}
    
    \{few\_shot\_examples\}
    
    **User Task Description (Includes data location information):**
    \{coding\_idea\}
    
    **Data Audit Preview (Includes original column names and sampled values):**
    \{raw\_preview\}
    
    **Output Requirements:**
    1. You must output exactly one complete JSON code block (wrapped in ```json).
    2. The JSON must contain the `uscp\_mapping` core field, strictly reflecting the correspondence between original columns and standard columns.
    3. The mapping logic must be based on the actual column names in the audit preview. If there is no corresponding column, use "None" instead.
    4. **CRITICAL: Rules for distinguishing cell\_type and cell\_line**:
       - **cell\_type**: Must map to the column describing the **cell's functional identity** (e.g., "T cell", "Neuron", "Hepatocyte").
       - **cell\_line**: If present, it must be mapped to `donor\_id` (e.g., "K562", "A549", "HEK293").
       - **Confused mapping prohibited**: If the data contains a `cell\_line` column and its values (e.g., "A549") also appear in the `cell\_type` column, this is considered an **incorrect mapping**.
       - **Correct Approach**:
         - If the data has both `cell\_type` and `cell\_line` columns -> Map `cell\_type` to the functional identity column and `donor\_id` to the `cell\_line` column.
         - If the data only has a `cell\_line` column (single cell line) -> Both `cell\_type` and `donor\_id` can map to `cell\_line`, but this must be explained in the `data\_summary`.
         - If the data only has a `cell\_type` column (primary cells) -> Map `cell\_type` to that column and `donor\_id` to other donor identifiers or "unknown".
    5. **Requirements for logical expressions**:
       - `is\_control\_logic` must be a valid Python boolean expression, such as `adata.obs['col'] == 'control'` or `adata.obs['col'].isin(['Ctrl', 'DMSO'])`.
       - `condition\_name\_logic` must be a valid Python string expression, such as `adata.obs['A'].astype(str) + '\_' + adata.obs['B'].astype(str)`.
    
    **JSON Structure Template:**
    \{ \{
      "uscp\_mapping": \{ \{
        "obs": \{ \{
          "cell\_type": "Original Column Name",
          "batch\_id": "Original Column Name or 'unknown'",
          "donor\_id": "Original Column Name or 'unknown'",
          "pert\_type": "Standard Value (drug/crispr/mixed/control) **Note: This is a constant value, not a column name**",
          "is\_control\_logic": "Python Boolean Expression",
          "condition\_name\_logic": "Python String Expression"
        \} \},
        "obsm": \{ \{
          "pert\_mask\_source": "Original Column Name",
          "pert\_dose\_source": "Original Column Name or 'None'"
        \} \},
        "var": \{ \{
          "index\_type": "Ensembl ID or Gene Symbol",
          "gene\_symbol\_col": "Original Column Name or 'None'"
        \} \},
        "numerical": \{ \{
          "is\_already\_log1p": true/false,
          "normalization\_required": true/false,
          "target\_sum": 10000.0
        \} \},
      "data\_summary": "A brief summary of the dataset (1-2 sentences)."
    \} \}}
\end{promptbox}

\begin{promptbox}[Prompt for Unifier Illustration]
    
    \agentrole{Universal Single-Cell Perturbation Dataset Standard Core Specifications:
    **\{Standard Field List\}:**
    - \{obs\}: `cell\_type`, `batch\_id`, `donor\_id`, `pert\_type`, `is\_control`, `condition\_name`
    - \{obsm\}: `pert\_mask` (Multi-hot encoding), `pert\_dose` (Normalized dose matrix)
    - \{var\}: Index must be \{Ensembl ID\}, including a `gene\_symbol` column
    - \{X\}: Must perform \{NormalizeTotal(target\_sum=1e4)\} and \{Log1p transformation\}
    
    **\{Technical Keys\}:**
    - After setting `var.index`, any redundant columns should be removed: `adata.var.drop(columns=['ensembl\_id'], inplace=True, errors='ignore')`}
\end{promptbox}

\begin{promptbox}[Prompt for Data Analysis]
    
    \agentrole{Task: Generate Deep Audit and Standardization Assessment Report for H5AD Datasets

    Data Paths:
    - Input H5AD Path: \{ h5ad\_path\_abs \}
    - Output Data Directory Path: \{ output\_data\_dir \}
    
    Requirements:
    1. **Deep Audit**: Load H5AD and, following the **USCP-DS v1.0** standard, identify potential corresponding columns for existing `batch\_id`, `cell\_type`, `pert\_type`, etc.
    2. **Semantic Mapping**: Analyze the value sampling of `obs` fields to determine the logical judgment for `is\_control` (e.g., which values represent the control group) and identify source columns for perturbation names and dosages.
    3. **Numerical Fingerprinting**: Check the statistical properties of the `X` matrix (min, max, mean) to clearly determine whether the data are raw counts or have completed Log-normalization.
    4. **Output Report**: Generate `data\_analysis\_report.md`, focusing on the "USCP-DS Mapping Recommendation Table" which lists the correspondence between original fields and standard fields.
    
    Constraints:
    - Adhere to the USCP-DS v1.0 specification framework.
    - **The standardized dataset after preprocessing (H5AD and .npz) must be saved to the aforementioned Output Data Directory**.
    - All paths must use absolute paths.}

\end{promptbox}

\begin{promptbox}[Prompt for Model Training]
    
    \agentrole{
    Role: Single-Cell Modeling and Deep Learning Architecture Expert
    
    Your task is to explore the optimal model architecture based on preprocessed standard datasets.
    
    0. Data Location - Must Use
    - **Training Set Path**: `\{path\_to\_train\_dataset\_pkl\}` (This absolute path must be used)
    - **Test Set Path**: `\{path\_to\_train\_test\_dataset\_pkl\}` (This absolute path must be used)
    - **Code Requirements**: Use `os.path.exists()` to verify the paths before loading. You must use the absolute path strings provided above directly without modification.

    1. Mandatory Model Interface Spec
    All implemented model classes must follow the following `forward` signature:

    def forward(self, x\_ctrl, pert\_dict):
        x\_ctrl: (Batch, Gene) - Individual cell expression of the control group
        pert\_dict: \{\{'mask': (Batch, N)\}\} - One-hot encoded perturbation mask
        return x\_pred: (Batch, Gene) - Predicted post-perturbation expression
        ...

    }
    2. MCTS Hierarchical Exploration Guidance
    Based on the Action selected by the current MCTS node, you need to choose one of the following implementation paths:
    
    * **Strategy: Discriminative Path**
    * Core Logic: `x\_pred = x\_ctrl + MLP\_Encoder(x\_ctrl, pert\_onehot)`
    * Available Modules: `models.blocks.PathwayAwareLayer` (used for gene alignment)

    * **Strategy: Generative Path**
    * Core Logic: `x\_pred = FlowMatchingEngine.solve(x\_ctrl, condition=pert\_dict)`
    * Available Modules: `models.blocks.FlowMatchingEngine`

    3. Data and Training Specifications
    
    * **CUDA Device Usage (Mandatory)**: All models, input data, and intermediate tensors must use `device = torch.device("cuda" if torch.cuda.is\_available() else "cpu")` and be moved to the CUDA device for execution.
    * **Data Loading and Parameter Probing (Sequence is Critical)**:
    1. **Must use the provided absolute paths**: `train\_path = "\{path\_to\_train\_dataset\_pkl\}"` and `test\_path = "\{path\_to\_train\_test\_dataset\_pkl\}"`
    2. **Path Verification**: Execute `assert os.path.exists(train\_path), f"Training set does not exist: \{train\_path\}"` before loading.
    3. Load the object using `dataset = PerturbationDataset.load(train\_path)`.
    4. **Model Parameter Acquisition**:
    * `input\_dim = dataset.X.shape[1]`
    * `num\_perts = dataset.pert\_mask.shape[1]` (i.e., total number of perturbation types)

    4. **Dataset Attributes**: `PerturbationDataset` contains `cell\_perturbations`, `cell\_condition\_keys` attributes, and the `get\_reference\_pool()` method. `MetricCalculatorTool` retrieves the global reference pool via `dataset.get\_reference\_pool()` and extracts metadata from `dataset.samples`.
    
    * **Data Format Consistency**:
    * **Unified Training and Testing Modes**: The dataset returns individual cell expressions in both training and testing modes, ensuring that training loss and validation loss are comparable.
    * **Fixed Control Group Pairing**: Each perturbed cell has a fixed control group pair, conforming to the control design principles of biological experiments.

    * **Perturbation Information Encoding (One-Hot + Learnable Embedding)**:
    * **One-Hot Encoding**: Use `pert\_dict['mask']` as a one-hot vector (shape: Batch * num\_perts) to indicate which perturbations are applied.
    * **Learnable Embedding Matrix**: Define `self.pert\_embedding = nn.Parameter(torch.randn(latent\_dim, num\_perts))` to convert one-hot vectors into embeddings.
    * **Implementation**: `pert\_repr = (self.pert\_embedding \@ pert\_dict['mask'].T).T` (shape: Batch * latent\_dim)

    * **DataLoader Configuration**: Ensure `drop\_last=True` is set (for the training set) to prevent BatchNorm from reporting errors on the final batch if its size is 1.
    * **Training Epochs**: Default training is **100 epochs**, unless Early Stopping is triggered.
    * **Data State**: Data is already log-normalized; **performing log1p again is strictly prohibited**.
    * **Machine Learning Priors (Sparsity Priors \& Distribution-Level Modeling)**:
    1. **Residual Learning**:

    * **Architectural Constraint**: Must adopt the `x\_pred = x\_ctrl + delta` structure.
    * **Biological Prior**: Gene regulatory networks exhibit sparsity. The model should learn the "Shift Vector" from the control state to the perturbed state, rather than re-generating the entire expression profile.

    2. **Loss Function**:

    * **Use MSE Loss**: Use `MSELoss` to calculate the error between predicted and ground truth values.
    * **Calculation Method**:
    * `x\_pred = x\_ctrl + delta\_pred` (Predicted post-perturbation state)
    * `x\_true = x\_post` (Ground truth post-perturbation state sample batch)
    * `loss = mse\_loss(x\_pred, x\_true)` (Directly calculate the Mean Squared Error between predicted and true values)

    * **Loss Function Formula**: `loss = mse\_loss(x\_pred, x\_post)`
    * **Biological Significance**: Through point-to-point error calculation, ensure the model accurately predicts the gene expression level of each cell after perturbation.
    * **Evaluation Metrics**: Evaluate using `tools.metric\_calculator.MetricCalculatorTool`. Returns 7 metrics: `RMSE\_Mean`, `Mean\_rank\_rmse`, `Cosine\_LogFC`, `LogFC\_rank\_cosine`, `PCC\_Delta`, `LogFC\_rank\_pcc`, `Matrix\_Distance`. Access via `res['metrics']['metric\_name']` after checking `res['success']
\end{promptbox}

\section{Limitations}
Despite \agentname's effectiveness, it faces limitations inherent to search-based systems. The MCTS engine entails higher computational overhead than static baselines, and the agent's creativity is currently bounded by a pre-defined library of architectural primitives. Furthermore, our framework focuses on unimodal data, leaving multi-modal integration and open-ended mathematical discovery as key directions for future research.

\end{document}